\documentclass[a4paper]{article}
\usepackage{INTERSPEECH2022}
\usepackage[dvipsnames]{xcolor}
\usepackage{amsmath,graphicx}
\usepackage{adjustbox}
\usepackage{multirow}
\usepackage{float}
\usepackage{booktabs}
\usepackage{array}
\usepackage{url}
\usepackage{algorithm}
\usepackage{algpseudocode}
\usepackage{enumitem}
\usepackage{cite}
\usepackage{hyperref}
\usepackage{cleveref}
\usepackage{multirow}
\usepackage{comment}
\usepackage{makecell}
\graphicspath{{'figures/'}}

\title{A two-stage transliteration approach to improve performance of a multilingual ASR}
\name{Rohit Kumar}
\address{NIT Patna, India}
\email{rohit1404060@nitp.ac.in}

\begin{document}

\maketitle

\begin{abstract}
End-to-end Automatic Speech Recognition (ASR) systems are rapidly claiming to become state-of-art over other modeling methods. Several techniques have been introduced to improve their ability to handle multiple languages. However, due to variation in writing scripts for different languages, while decoding acoustically similar units, they do not always map to an appropriate grapheme in the target language. This restricts the scalability and adaptability of the model while dealing with multiple languages in code-mixing scenarios. This paper presents an approach to build a language-agnostic end-to-end model trained on a grapheme set obtained by projecting the multilingual grapheme data to the script of a more generic target language. This approach saves the acoustic model from retraining to span over a larger space and can easily be extended to multiple languages. A two-stage transliteration process realizes this approach and proves to minimize speech-class confusion. We performed experiments with an end-to-end multilingual speech recognition system for two Indic Languages, namely Nepali and Telugu. The original grapheme space of these languages is projected to the Devanagari script. We achieved a relative reduction of 20\% in the Word Error Rate (WER) and 24\% in the Character Error Rate (CER) in the transliterated space, over other language-dependent modeling methods.
\end{abstract}

\noindent\textbf{Index Terms}: Speech recognition, Transliteration, Code-Mixing, Language-agnostic, multilingual model

\section{Introduction}
Automatic Speech Recognition (ASR) systems have experienced a paradigm shift in recent years, from statistical modeling towards end-to-end solutions. Traditional ASR systems relied on appropriate parametric representation to train acoustic models and aided with optimal pronunciation lexicon and language models, dealing with finding the best sequence of words (W)\cite{povey2011kaldi}. For such systems, acoustic models are realized over good quality audio recordings, aided with the corresponding transcriptions, by appropriately mapping the statistical characteristics of the audio segment to the label. The language model (LM) helps derive a valid sequence of words once the decoding has been done\cite{picone1990continuous}.

End-to-end ASR systems directly map the sequence of input acoustic features to a sequence of graphemes or words. The system encapsulates speech recognition components such as the acoustic, language, and lexicon model into a single pipeline, thus bypassing the required linguistic resources while transferring the useful bias from the source model to the target language\cite{graves2012sequence, graves2014towards, chan2016listen, prabhavalkar2017comparison, chiu2018state}.
Usually, the performance of end-to-end ASR systems over traditional systems has an absolute edge. For popular languages, these systems have already been able to handle gender, dialect, and age diversity. However, for scenarios dealing either with low resources or unseen languages, the performance of these systems suffers considerably. Such scenarios include code-switching, code-mixing, and low resource cases. Without an explicit language specification, the model usually confuses between multiple languages, especially for languages that have very little overlap in their scripts.

India is a diverse country comprising multicultural societies, hosting a hub of several bilingual speakers. Indic languages exhibit significant diversity, and this creates challenges in realizing a good performing ASR system. Different languages have their native script, and furthermore, there is considerable overlap of acoustic content across languages. Hence it is quite likely for acoustically similar segments to be mapped to different graphemes in any ASR system. In multicultural societies, overlapping of language information such as common words, lexical content, occurs quite often, due to language family relations and the geographical significance of the speakers. In India, there are 22 official languages, each with quite a variety of dialects. For several speakers, multiple language pairs frequently appear, mostly in combination with English (code-mixing)\cite{bhuvanagiri2010approach}. Code-mixing appears as a word or clausal level embedding of one language in the matrix of another and poses quite a challenge for ASR systems to handle. In India, code-mixing with English is an inherent characteristic for speakers of several regions, for example, Telugu-English in Southern, Nepali-English in North-Eastern, and Hindi-English in Northern parts of India. 

Multilingual models for speech recognition have evolved from the early Hidden Markov Model-based to Deep Neural network-based architectures, and even for the recent state-of-art end-to-end systems\cite{thomas2012multilingual, tuske2013investigation, sercu2017network}. Generally, speech recognition of multiple languages requires a lot of training data with the corresponding annotation, hence it's difficult to train the acoustic model for low resource languages. Acoustic models trained in one language, when borrowed do not perform well in the context of other languages. The reason for this is the phonetic content mismatch between the source language and the target language. These factors limit the performance of speech recognition modules in the context of multilingual scenarios.

In this paper, we propose a two-stage transliteration approach that helps the model learn internal representation, thus enhancing acoustic-phonetic content from languages. Transliteration is a way of representing how the phonetic content of words of one language can be represented in the lexicon of another language. Transliteration is mostly employed for tasks such as machine translation, text retrieval, and ASR systems. For a source language, transliteration provides a unique set of characters in a target grapheme space, thus allowing the mapping of acoustically similar units to a single sequence of graphemes. Apart from such acoustically similar units, this mapping is usually bijective in nature, and hence the transliterated text can easily be mapped to the target language domain in an error-free manner. For units similar in nature, language models in the target language can help determine the appropriate grapheme in case of confusion. The proposed two-stage transliteration prioritizes the reduction of grapheme scope for reduced confusion during the phoneme-grapheme mapping, hence enhancing the performance of ASR in the transliterated space. Organization of the paper is as follows: Section 2 describes the related work done, Section 3 describes the proposed transliteration approach, Section 4 briefs the data, experimental setup, and result analysis, and is followed by the conclusion in Section 5.

\section{Related work}

For end-to-end ASR systems, MFCC features extracted from the time-domain signal serve as a popular choice for the neural network\cite{hannun2014deep}. The neural network maps the input frame to grapheme space in the target language. A vocabulary accommodating different characters from multilingual data induces speech-class ambiguity for acoustically similar units. Most of the recent ASRs have been developed over large datasets, with over 10,000 hours of audio recordings, and even use data augmentation methods over high-resource language to realize sophisticated systems\cite{amodei2016deep}.

\cite{pandey2017adapting} proposed an approach where monolingual Hindi acoustic models were trained through phoneme adaptation. Often in recognition of code-mixed/switched speech, mixing/switching or borrowing of words is mostly confusing due to the lack of standardization of code-mixed transcripts leading to the presence of homophones. \cite{srivastava2018homophone} proposed the use of a WX-based standard pronunciation scheme to automatically identify and disambiguate homophones.
 
Multilingual modeling refers to training over multiple languages for a single network pipeline and recognizing utterances corresponding to the diverse language set. In previous works, \cite{ward1998towards, wang2002towards} combined different multilingual acoustic models and language models (LM) together in a probabilistic framework. \cite{fugen2003efficient} combined different monolingual LMs into multilingual LM and introduced the concept of grammar into multilingual LM. The set of shared units, the structure of shared states, and the shared acoustic-phonetic properties help the language universal and language adaptive acoustic model to outperform the language-specific model\cite{lin2009study}. Multilingual acoustic models have subsequently outperformed monolingual acoustic models\cite{heigold2013multilingual}. Recently, a grapheme-based sequence-to-sequence model is trained by taking a union of language-specific grapheme sets on different Indian languages together\cite{toshniwal2018multilingual}. While training high resource language with low resource language, the performance of high resource language degrades due to heterogeneous multilingual data\cite{li2021scaling}.
 
Transliteration has been widely used with different applications to write words from one language into another language. Transliteration has also been used as a metric for evaluating the modeling errors in the acoustic and language model\cite{emond2018transliteration}. Transliteration helps the acoustic model to reduce the number of modeling units, which eventually leads to a more flexible method for a newer language.

\section{Proposed Transliteration Approach}
\label{sec:Proposed approach}
For code-mixing/code-switching scenarios, similar phonetic entities align themselves with different graphemes, according to their writing script, which leads to phonetic confusion while creating a super-set of like-sounded phones during training. This section introduces the proposed two-stage transliteration approach to handle multiple languages in an end-to-end schema. 

The proposed two-stage Phoneme-to-Grapheme based transliteration approach is based on mapping the phoneme set of the target language to intermediate language grapheme sets. The target language ($l_{tgt}$) here is the language, or a mix of multiple languages that serve as input to the ASR. The intermediate language ($l_{int}$) is the language whose script is being transformed to. Indic languages have a complex combination of consonants, vowels, and diacritics, which makes it difficult to map to Roman script, and restricts the encoding of language information. The challenges to map Indic languages with other Latin scripts have been highlighted in literature\cite{emond2018transliteration}. The proposed approach hence chooses the Devanagari script as $l_{int}$ to convert the target languages $l_{tgt}$ Nepali, and Telugu. These languages share some common phonemes with the Devanagari script and hence the transformation is smooth. Following are the two stages of the proposed transliteration approach. 

\subsection{Stage-1 : Lexical Transformation}
\begin{table}
    \centering
    \includegraphics[width=5cm,height=8cm]{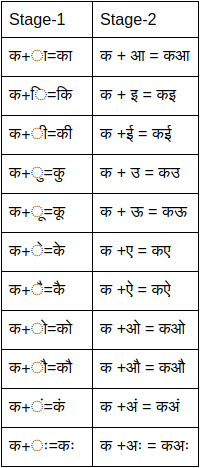}
    \caption{Representation of characters in the two-stage transliteration method.}
    \label{tab:my_label1}
\end{table}
\label{ssec:subsubhead3}
Stage-1 addresses the lexical representation of $l_{tgt}$ to script of $l_{int}$. A lexicon of the Nepali and Telugu graphemes is formed containing all the native and foreign words (which constitute the code-switching/mixing scenario). A mapping is used to convert all the graphemes to their corresponding phoneme sequence. The phonemic sequence of $l_{tgt}$ is then mapped to the corresponding phonemic sequences of the $l_{int}$ Devanagari script. Phonemic sequences in $l_{int}$ are then arranged according to vowel-consonant rules for word formation according to the Devanagari script. Finally, the phonemic sequence in $l_{int}$ Devanagari script are mapped to their corresponding grapheme sequence. For $l_{tgt}$ Nepali, combined words are split initially into individual words to avoid ambiguity and to enrich the lexicon.

\subsection{Stage-2 of transliteration}
\label{ssec:subsubhead}
After Stage-1, a major issue of mapping acoustically similar entities to a common sequence of graphemes remains. Stage-2 addresses this by focusing on converting the dependent form (\textit{matra}) of a vowel to the independent form. This conversion allows the same phonetic sound of \textit{matras} to get mapped to graphemes of their independent form of vowels. This mapping minimizes the speech-class confusion as well by limiting the size of the vocabulary. Table~\ref{tab:my_label1} shows this conversion of \textit{matra} to its independent form of vowel. Table~\ref{tab:my_label2} illustrates the 2-stage process with different examples for both target languages. 

\section{Data, Experiment and Result Analysis}
\label{sec:evaluation}

\subsection{Data}
\label{ssec:subhead}
We conducted experiments on data from 2 target languages, namely Nepali and Telugu, which overall corresponds to the total amount of 160 hours of training, and 26 hours of test data. Nepali data is collected from read-out phrases by native speakers\cite{kjartansson-etal-sltu2018} and contributes to 113 hours of training data. Telugu data is collected from read-out phrases and conversational speech\cite{srivastava2018interspeech} and contributes to 47 hours of training data. The data are recorded at a sampling rate of 16 kHz with a 16-bit resolution. Both languages have been manually transcribed into their native grapheme set. Both training and testing data exhibit a considerable proportion of code-mixing with English. 

\begin{table}
    \centering
    \includegraphics[width=6cm,height=5cm]{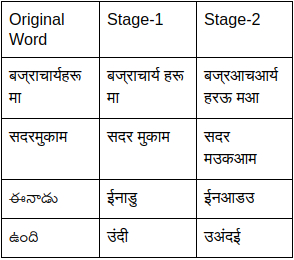}
    \caption{Illustration of examples using a two-stage transliteration method.}
    \label{tab:my_label2}
\end{table}

\subsection{Experimental Setup}
\label{ssec:subhead1}
The method uses an end-to-end deep-learning architecture with connectionist temporal classification (CTC) loss function as proposed in Deepspeech2\cite{deepspeech2} for building ASR system. The network composes of 2 CNN layers followed by 5 RNN layers, and a softmax layer with CTC Loss criterion. The system is usually trained on inputs as audio segment or equivalent frequency-domain representation, with the expected output being the corresponding grapheme representing the phonetic content. Model parameters and feature specifications have slightly been modified for tuning the model to a multilingual scenario. In the current work, spectrogram of power normalized audio segments are used as input features. We considered 96 audio features derived over a window size of 20 ms, at a stride of 10 ms, transformed over 512 DFT bins, at an audio sampling rate of 16 KHz. Beam search decoder with language model scoring and beam width of 512 is used. Labels from the proposed transliteration scheme represent the output.

\begin{table}[]
\centering
\begin{tabular}{|l|l|l|l|l|l|l|}
\hline
\multirow{2}{*}{Experiment} & \multicolumn{2}{l|}{Nepali} & \multicolumn{2}{l|}{Telugu} & \multicolumn{2}{l|}{Average} \\ \cline{2-7} 
                & WER   & CER   & WER   & CER   & WER   & CER   \\ \hline
E0(Baseline)    & 32.1  & 10.4  & 47.6  & 22.3  & 39.8  & 16.3  \\ \hline
E1              & 30.7  & 9.0   & 45.3  & 20.2  & 38    & 14.6  \\ \hline
E2 (stage-1)    & 25.4  & 8.3   & 41.8  & 18.1  & 33.9  & 13.9  \\ \hline
E3 (stage-2)    & 23.8  & 7.8   & 39.5  & 15.7  & 31.6  & 11.7  \\ \hline
\end{tabular}
\caption{Performance (WER/CER) of ASR system after Stage-1 and Stage-2 of transliteration.}
\label{tab:my-result}
\end{table}

\subsection{Result Analysis}
\label{ssec:subhead2}
In this section, we discuss the effect of our proposed two-stage phoneme-to-grapheme based transliteration approach over the multilingual acoustic model. In Table~\ref{tab:my-result}, we present the performance of the ASR system after the different stages of transliteration. The baseline for the ASR system is represented as experiment E0. Experiment E1 represents training the acoustic model without transliteration. Experiment E2 represents training the acoustic model with stage-1 of the proposed transliteration approach, and experiment E3 represents training the acoustic model with stage-2 of the proposed method. A significant improvement in the performance of the ASR system after each stage of transliteration concerning that of the acoustic model trained without transliteration can be noted in the Tab.~\ref{tab:my-result}. After stage-1, the language information among the different languages is encoded into a single writing script, resulting in reduction in number of classes during the phone classification. And in Stage-2, after the replacement of matra with its full vowel, the vocabulary size gets further reduced, having a minimal number of characters left. Thus, reducing the speech-class confusion and allowing mapping of similar-sounding acoustics to individual graphemes. We observe a reduction of 22.4

\section{Conclusion}
\label{sec:conclusion}

In this paper, we explored a transliteration-based approach to address the code-mixing scenario in a monolingual acoustic model. Our motivation is to improve upon the performance while avoiding retraining the acoustic model, which is usually done in case of overlapping of language information among multiple languages. We propose a two-stage phoneme-to-grapheme based transliteration approach, while choosing a popular and standard language as intermediate script for multiple languages in the target set. The proposed transliteration method operates in two stages. In the former stage, we perform transliteration with the dependent vowels, i.e., matra. In the latter stage, we map the dependent form of a vowel to its independent form. We achieve an average reduction of 20

\bibliographystyle{IEEEtran}
\bibliography{strings,refs}

\end{document}